\documentclass[11pt]{article}
\usepackage{eamt15}
\usepackage{times}
\usepackage{latexsym}
\usepackage{graphicx}
\usepackage[hyphens]{url}

\makeatletter
\newcommand{\@BIBLABEL}{\@emptybiblabel}
\newcommand{\@emptybiblabel}[1]{}
\makeatother
\usepackage{hyperref}

\title{Content Translation: Computer-assisted translation tool \\ for Wikipedia articles}
\author{Niklas Laxstr\"om\\
  University of Helsinki\\
  Dept. of Modern Languages\\
  and \\
  Wikimedia Foundation\\
  {\small {\tt nlaxstrom@wikimedia.org}}  \And
  Pau Giner\\
  Wikimedia Foundation\\
  {\small {\tt pginer@wikimedia.org}}  \And
  Santhosh Thottingal\\
  Wikimedia Foundation\\
  {\small {\tt sthottingal@wikimedia.org}}}

\date{}

\begin{document}

\maketitle

\begin{abstract}

The quality and quantity of articles in each Wikipedia language varies greatly. Translating from another Wikipedia is a natural way to add more content, but the translation process is not properly supported in the software used by Wikipedia. Past computer-assisted translation tools built for Wikipedia are not commonly used. We created a tool that adapts to the specific needs of an open community and to the kind of content in Wikipedia. Qualitative and quantitative data indicates that the new tool helps users translate articles easier and faster.
\end{abstract}

\section{Introduction}

Wikipedia is the most multilingual encyclopedic knowledge archive, with over 280 languages with varying amount of content. Knowledge available for a user is limited by the languages used to access it. Translation has been a common way to expand knowledge across languages in Wikipedia. The editing activity of the top 46 language editions of Wikipedia shows that 25\% of edits by multilingual users are for the same article in different languages~\cite{Hale13}.

It is not necessary to use any tool to translate Wikipedia articles. However, it is a complicated process and mainly done by experienced Wikipedia editors.


There were many attempts to build tools to support translation of articles. 
None has seen widespread use:
in our research only few users reported using those tools when translating Wikipedia articles.

In this paper we present a new approach to support translation which has been designed taking into account the unique needs of Wikipedia content
and their community.
Content Translation (CX) is a new tool that automates many steps of the translation process and validates the approach in practice. 
It was first enabled on 8 Wikipedias as an opt-in feature to create new articles in January 2015. Selected language pairs have machine translation (MT) support.




\section{Previous work}
\label{previous-work}

MediaWiki, the software powering Wikipedia, is translated to hundreds of languages using the Translate extension. No such solution was available for translating Wikipedia articles, leaving a gap in the translation support.

There were at least ten instances of translation tools built for Wikipedia\footnote{Details collected at \url{https://meta.wikimedia.org/wiki/Machine_translation}}. Those tools can be divided into two groups based on whether the tool creators already possessed MT software. The first group is composed of companies such as Google and Microsoft, but also smaller companies and researchers. The other group of tools has been created just for Wikipedia article translation, mostly by volunteers.


Among the earliest tools were GTT by Google and WikiBhasha by Microsoft, using their own MT services~\cite{garcia2009reviews,kumarana2011wikibhasha}. Later, Casmacat for professionals and researchers~\cite{alabau2013casmacat} and CoSyne for multilingual MediaWikis~\cite{Bronner:2012:CSM:2462932.2462973}, unlike Wikipedias which are monolingual.



Common to all such tools is that they are not integrated into Wikipedia. To use them one needs to go to another website or install software. CX is integrated into Wikipedia and provides a WYSIWYG editor (what you see is what you get).

\section{Designing the translation experience}

The design of CX was aimed at improving the existing process users followed when translating. Following the principles of User-Centered Design~\cite{Norman86}, we organised periodic user research sessions to (a) better understand the user needs during the existing translation process, and (b) validate new ideas on how to improve this process. 

\subsection{User research}


We recruited 106 participants using a survey\footnote{\url{https://goo.gl/iKQIDh}}. From their responses we identified dictionaries (76\% of participants used them), and Wikipedia (60\%) as their most used tools when translating. MT (53\%), spell checkers (48\%) and glossaries (42\%) were also common. Less than 6\% of the participants mentioned tools specifically aimed at Wikipedia article translation, such as those described in Section~\ref{previous-work}, and no tool was mentioned by more than one participant.

We organised 16 research sessions. Sessions were organised in two parts. In the first part, contextual inquiry techniques~\cite{Beyer98} were applied to observe user behaviour while translating, and identify their needs. The second part was a usability testing study~\cite{Jakob94} to evaluate different design ideas in the form of prototypes.



\subsection{Design principles}
\label{principles}

\begin{figure*}[ht]
\centering
\includegraphics[width=1\textwidth]{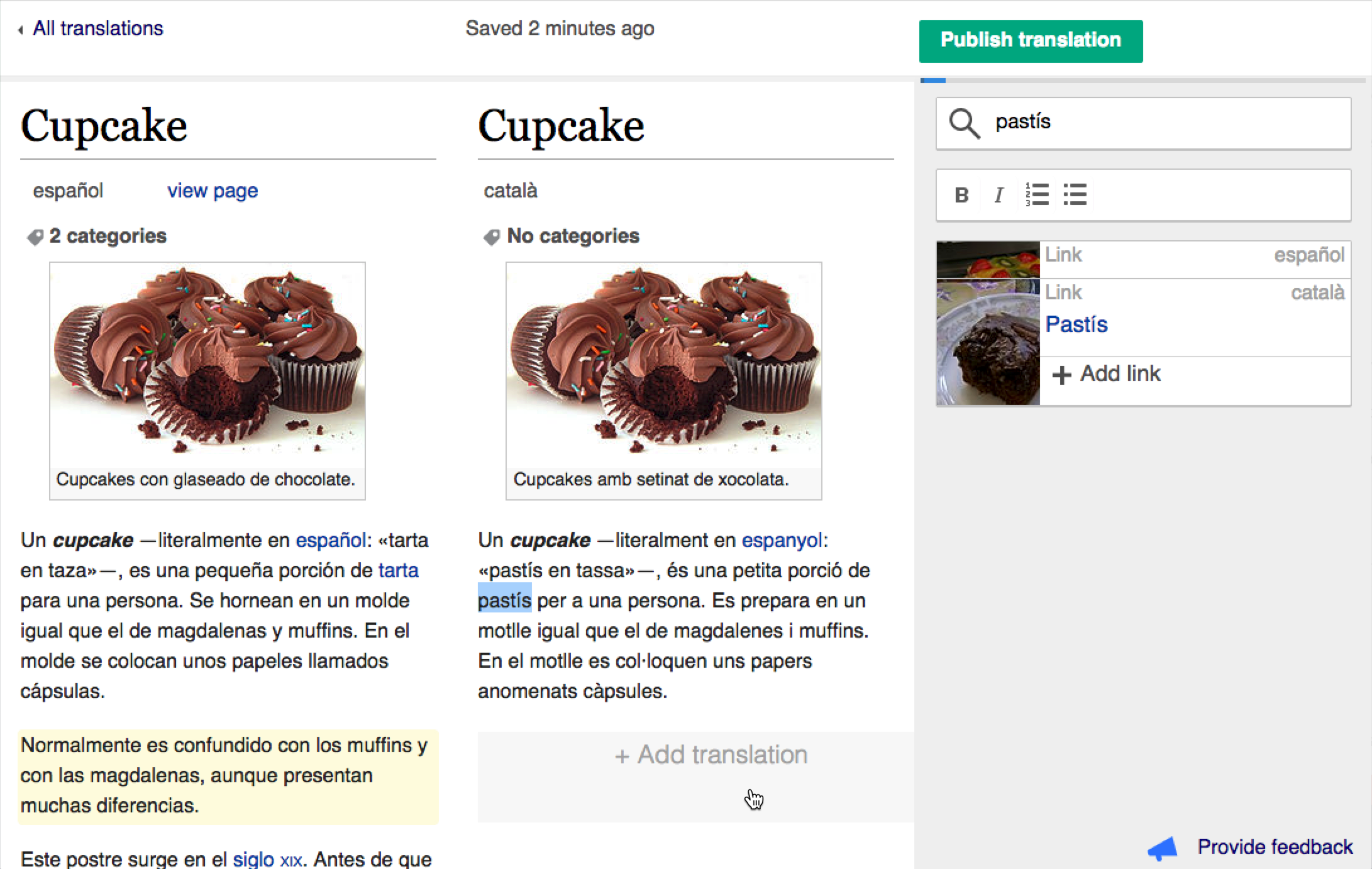}
\caption{The source and translated content side-by-side and additional tools on the right. \label{cupcake}}
\end{figure*}

The research sessions were instrumental to guide the design of the translation experience\footnote{A detailed design specification is available at~\url{https://www.mediawiki.org/wiki/CX}}. The following design principles summarise the approach we followed when designing the tool.


\subsubsection{Freedom of translation}

There is a significant diversity in Wikipedia content across languages. On average, two articles from different languages on the same topic have just 41\% of common content~\cite{Hecht10}. In contrast to other kinds of content, such as software user interface strings or documentation, Wikipedia articles are not intended to be exact translations that are always kept in sync. In order to support that content diversity, CX does not force users to translate the full article. As illustrated in Figure~\ref{cupcake}, users add content to the translation one paragraph at a time. When a paragraph is added, an initial translation based on MT is provided for the user to edit. MT is used if available, but the user can also start with the source text or an empty paragraph if that is preferred.

Unlike other tools that define a strong boundary to translate on a per sentence basis, working at a paragraph level allows users to reorganise sentences and accommodate different editing patterns.

\subsubsection{Provide context information}
    In CX the original article and the translation are shown side-by-side. Each paragraph is dynamically aligned vertically with the corresponding translated paragraph, regardless of the difference in length. This allows users to quickly have an overview of what has already been translated and what has not.

    Contextual information reduces the need for the user to navigate and reorient. When translating a sentence, the corresponding sentence in the original document is highlighted. In addition, when manipulating the content, options are provided anticipating the user's next steps. In Figure~\ref{cupcake}, based on the user's text selection, the user can explore the article related to the selected text (in the source or target languages), or turn the selected text into a link. Dictionary can be accessed inside the tool by selecting a word or by using the search box in the tools column.

\subsubsection{Focus on the translation}

We identified many steps in the translation process that could be automated. Users spend time making sure each link they translated points to the correct article in the target Wikipedia, and recreating the text formatting that was lost when using an external translation service. They also look for categories available in the target Wikipedia to classify the translated article, and save constantly during the process to avoid losing their work.

CX deals with those aspects automatically. When adding a paragraph, the initial translation preserves the text format. Modifications to the translated content are saved automatically. Links point to the right articles if they exist and existing categories are added to the article thanks to Wikidata\footnote{\url{https://www.wikidata.org}}, a structured data knowledge repository, that maps corresponding concepts across languages. As those aspects are automated, users can focus on adapting content for the initial version of the article rather than on technical and formatting tasks.


 
\subsubsection{Quality is key}

One of the concerns raised early by the participants was about MT quality. Users were concerned about the potential proliferation of low quality content in Wikipedia articles.

    In order to respond to that concern, CX keeps track of the amount of text that is added from MT without further modification by the users. When a given threshold is exceeded, a warning is shown to users encouraging them to focus on quality more than quantity. 


\section{WYSIWYG implementation}
MediaWiki's wikitext is not standardised. For a long time, the only way to use wikitext was to render it to HTML with MediaWiki. Parsoid\footnote{\url{https://www.mediawiki.org/wiki/Parsoid}} is a Wikimedia project that implements a second parser for wikitext. To follow the principle \emph{focus on translation} principle
we only provide limited editing and formatting options and side-step a lot of complexity of Wikipedia article structure without negatively affecting the translation process. CX is the first translation tool that provides a WYSIWYG editor using the annotated HTML provided by Parsoid.

Some MT services neither support HTML input nor provide reordering information. Preserving markup is an essential requirement for CX because wikitext adaptation and WYSIWYG editing are based on the markup. We devised an algorithm that can reconstruct the reordering information by making the MT service do some additional work.


\section{MT evaluation}
We use the subjective evaluations of MT quality for a given language pair to decide whether we will include a MT service for a language pair in the tool. To evaluate a MT for a given language pair, we ask the potential future users of the tool to translate articles using it and tell whether it was useful for them or not.

We run a MT service on our servers using the open-source Apertium project~\cite{forcada2011apertium}, but we support other MT providers as well. 

\section{Evaluation}

Currently the tool is only available to self-selected users (most of them experienced editors), hence the results cannot be generalised to the whole community. Further studies on the resulting quality over a long term will help.

The low deletion ratio for articles created using CX suggests that there are no major problems in terms of quality. In three months of exposing the tool as an opt-in feature, 900 articles were published using CX with an overall deletion ratio lower than 1\% across all languages, which is lower than the deletion rate for all new articles.



We noticed that there is a significant difference between the number of created articles in different target language Wikipedias, which cannot be explained by the number of active users, number of available articles to translate nor the availability of MT.
For example in three months the Catalan Wikipedia saw 455 articles created by translating from Spanish with CX, but in the Portuguese Wikipedia only 25. Both language pairs have MT provided by Apertium.
Statistics about the tool are collected publicly\footnote{\url{https://www.mediawiki.org/wiki/Content_translation/analytics}}.

We have not yet made precise measurements on translation time saving, but we got positive reports from our users. In a roundtable\footnote{\url{https://blog.wikimedia.org/2014/09/29/round-table-with-editors-from-the-catalan-wikipedia/}} organised with editors of the Catalan Wikipedia, an experienced editor reported a 70\% time saving.

We found that English is the most used source language, consistent with Hale's findings on multilingual user behaviour~\shortcite{Hale13}.

\section{Conclusions}

We developed a tool that addresses the specific needs of an open community and the specifics of the kind of content in Wikipedia. CX is a computer-assisted translation tool with a WYSIWYG editor and automatic link adaptation. CX supports multiple different MT providers, but by integrating the open source Apertium project we were able to quickly provide MT for multiple language pair 
We developed MT education and tracking features to address community concerns about proliferation of poor quality translations.

User feedback for CX is supportive and data also shows that quality of the published translations is good, alleviating the community concerns. The low translation activity in multiple languages where the tool is already available needs further research. Close integration in Wikipedia allows CX to recruit users and suggest articles to translate in ways not possible with the previous tools.


\bibliography{eamt15}
\bibliographystyle{eamt15}
\end{document}